\documentclass{article}

\usepackage{arxiv}
\usepackage{float}
\usepackage[utf8]{inputenc}
\usepackage[T1]{fontenc}
\usepackage{hyperref}
\usepackage{url}
\usepackage{booktabs}
\usepackage{amsfonts}
\usepackage{amsmath}
\usepackage{nicefrac}
\usepackage{microtype}
\usepackage{graphicx}
\graphicspath{{./images/}}

\title{Cryptocurrency Price Prediction Using Parallel Gated Recurrent Units}

\author{
  Milad Asadpour \\
  College of Interdisciplinary Science and Technology\\
  University of Tehran\\
  Tehran, Iran\\
  \texttt{milad.asadpour@ut.ac.ir} \\
  \And
  Alireza Rezaee \\
  College of Interdisciplinary Science and Technology\\
  University of Tehran\\
  Tehran, Iran\\
  \texttt{arrezaee@ut.ac.ir} \\
  \And
  Farshid Hajati \\
  School of Science and Technology\\
  University of New England\\
  Armidale, NSW, Australia\\
  \texttt{fhajati@une.edu.au} \\
}

\begin{document}
\maketitle

\begin{abstract}
According to the advent of cryptocurrencies and Bitcoin, many investments and businesses are now conducted online through cryptocurrencies. Among them, Bitcoin uses blockchain technology to make transactions secure, transparent, traceable, and immutable. It also exhibits significant price fluctuations and performance, which has attracted substantial attention, especially in financial sectors. Consequently, a wide range of investors and individuals have turned to investing in the cryptocurrency market. 
One of the most important challenges in economics is price forecasting for future trades. Cryptocurrencies are no exception, and investors are looking for methods to predict prices; various theories and methods have been proposed in this field. This paper presents a new deep model, called \emph{Parallel Gated Recurrent Units} (PGRU), for cryptocurrency price prediction. In this model, recurrent neural networks forecast prices in a parallel and independent way. The parallel networks utilize different inputs, each representing distinct price-related features. Finally, the outputs of the parallel networks are combined by a neural network to forecast the future price of cryptocurrencies.
The experimental results indicate that the proposed model achieves mean absolute percentage errors (MAPE) of 3.243\% and 2.641\% for window lengths 20 and 15, respectively. Our method therefore attains higher accuracy and efficiency with fewer input data and lower computational cost compared to existing methods.
\end{abstract}


\section{Introduction}

The current economic system is based on intermediaries and third parties such as banks and financial institutions. These intermediaries are required for most financial transactions, but this reliance introduces drawbacks including limited trust, security, transparency, and flexibility. In 2009, Bitcoin was introduced to the public to facilitate financial transactions and address some of the shortcomings of the current financial system \cite{nakamoto2008bitcoin}.

Bitcoin does not exist physically; users hold digital accounts identified by public and private keys. Account information, along with transaction data, is recorded in a global ledger (the blockchain). Maintaining this ledger requires substantial computational power to verify and record data across the distributed network. Bitcoin is generated by solving computationally expensive mathematical problems, creating a direct peer-to-peer connection between individuals in each transaction and eliminating intermediaries and third parties. There are no central servers or financial institutions governing transfers.

A decade after its advent, Bitcoin has received widespread attention due to its increasing value and volatile performance. From January 2010 to January 2021, the lowest recorded price was \$0.1 in July 2010, and the highest was \$40,921 in January 2021. Over the same period, Bitcoin's market capitalization rose from around \$16.3 billion in February 2017 to nearly \$760 billion in January 2021. Although price movements appear highly volatile and possibly random, closer inspection (e.g., via logarithmic price diagrams) indicates the presence of recurrent patterns \cite{site_investing,site_bitcoin}.

These fluctuations and substantial increases in the price of cryptocurrencies have motivated many researchers to search for influential factors and robust methods for predicting Bitcoin prices. Various approaches, including technical and fundamental analysis in economics, have been used for this purpose \cite{patel2020deep,huang2019predicting,akyildirim2020volatility}. Machine-learning methods are among the most promising tools for predicting future cryptocurrency prices. They can generally be categorized into classification and regression approaches. In classification tasks, the goal is to predict a directional movement (e.g., increase or decrease) \cite{chen2020bitcoin,mcnally2018predicting}, whereas in regression tasks, the goal is to predict the actual price value.

Deep neural networks (DNN), recurrent neural networks (RNN), and convolutional neural networks (CNN) have been widely applied to both regression and classification problems in this domain. Similar advanced deep architectures have been successfully used in other domains such as face analysis \cite{Hajati2006FaceLocalization,Pakazad2006FaceDetection,Hajati2010PoseInvariant,Hajati2017DynamicTexture,Ayatollahi2015,Hajati2017Surface,AbdoliHajati2014,Shojaiee2014Palmprint,CremersACCV2014}, load forecasting \cite{Barzamini2012}, and medical decision support \cite{Fiorini2019,Mahajan2024,Sadeghi2024COVID,Sopo2021DeFungi,Sadeghi2024ECG,Tavakolian2022FastCOVID,Tavakolian2023Readmission,Wang2022SoftwareImpacts}.

In this work, we study and predict the price of Bitcoin as a regression problem using a new model called \emph{Parallel Recurrent Neural Network} (PRNN), instantiated via gated recurrent units (GRUs). Our approach is also motivated by prior work on evolutionary and learning-based optimization in control and communications \cite{Rezaee2010FIR,KarimiRezaee2017Helmholtz,MohamadzadeRezaee2017Antenna,RezaeeGolpayegani2012,Rezaee2017PID,Rezaee2017MPC,Rezaee2017Penetrometer,Rezaee2014FuzzyCloud,Gavagsaz2018LoadBalancing,Ramezani2024Drones} and in complex multi-agent and educational systems \cite{Sarvghad2011ThinkingStyles,Rezaee2015coevolutionary}.

The proposed structure comprises three small neural networks. The first and second networks are based on GRU layers and are arranged in parallel, operating independently. The first network predicts the Bitcoin price based on historical price features. The second network predicts the price based on structural features of the Bitcoin network. Finally, the outputs of these two networks are fed into a third network---a feedforward neural network---which combines them based on their respective errors and reliabilities to produce the final price prediction for the next few days.

By using different but complementary input data (price and structural features) in parallel networks, price fluctuations and patterns are captured more completely, yielding highly accurate and reliable forecasts. Our experimental results show that the proposed model attains MAPE of 3.243\% and 2.641\% for window lengths 20 and 15, respectively. Thus, the method uses fewer input features while achieving higher accuracy and lower computational cost compared to existing approaches.

\section{Related Work}

Numerous studies have aimed to identify accurate indicators and influential factors for Bitcoin prices and other digital currencies. Teker et al.\ \cite{teker2019determinants} investigated relationships between the prices of gold, crude oil, and major cryptocurrencies such as Bitcoin, Ethereum, and Litecoin. Their findings help explain cryptocurrency price fluctuations at the international level.

Peng et al.\ \cite{peng2018best} combined an autoregressive conditional heteroscedasticity model with support vector regression to predict the prices of Bitcoin and Litecoin, achieving better performance than autoregressive moving-average methods. Karasu et al.\ \cite{karasu2018prediction} used SVMs with linear and polynomial kernels, trained and evaluated via $k$-fold cross-validation, and concluded that SVMs generally outperform linear regression.

Hitam et al.\ \cite{hitam2019optimized} further improved SVM performance by optimizing its coefficients with particle swarm optimization (PSO), resulting in more accurate and relatively faster predictors. Radityo et al.\ \cite{radityo2017prediction} applied genetic algorithms to optimize the weights of a multilayer neural network, helping the model avoid local minima and improving training quality. Rathan et al.\ \cite{rathan2019crypto} proposed a hybrid approach combining decision trees and linear regression, achieving lower computational cost and slightly higher accuracy than some earlier methods.

More recently, researchers have focused on deep learning. Chen et al.\ \cite{chen2020bitcoin} used feature selection to identify twelve critical features affecting the Bitcoin price and then applied classification methods such as long short-term memory (LSTM) networks and quadratic discriminant analysis to predict future price direction (up or down). Their results showed that using these twelve features with LSTMs yields higher accuracy than other classification methods.

Wu et al.\ \cite{wu2018new} proposed an LSTM-based forecasting framework using time-series data of Bitcoin prices. Tandon et al.\ \cite{tandon2019bitcoin} used an LSTM network trained with $k$-fold cross-validation, significantly increasing computational cost but improving the model’s ability to capture complex price dynamics. Rizwan et al.\ \cite{rizwan2019bitcoin} compared several deep learning models and showed that GRU-based networks outperform LSTM-based models by approximately 10\% on average in terms of accuracy, while being computationally simpler. Their network used both price data and key cryptocurrency indicators such as total trading volume, average hashrate, and daily fees.

Beyond price and structural features, several works have incorporated sentiment and attention signals. Jain et al.\ \cite{jain2018forecasting} leveraged Twitter data and sentiment analysis to forecast Bitcoin prices. Smuts et al.\ (conceptually similar to \cite{mohanty2018predicting}) used Telegram messages and Google Trends data, finding that the volume of Google searches strongly impacts short-term cryptocurrency prices, especially Bitcoin. Mohanty et al.\ \cite{mohanty2018predicting} used an extensive set of 26 Bitcoin-related features---including price, daily transactions, number of blocks, block generation time, and Twitter data---to predict price fluctuations, achieving high accuracy but at a substantial computational cost due to the large input dimensionality.

In parallel, deep and machine-learning models have been applied to a broad range of forecasting and recognition problems: short-term load forecasting \cite{Barzamini2012}, medical imaging and pattern recognition \cite{Hajati2006FaceLocalization,Pakazad2006FaceDetection,Hajati2010PoseInvariant,Hajati2017DynamicTexture,Ayatollahi2015,Hajati2017Surface,AbdoliHajati2014,Shojaiee2014Palmprint,CremersACCV2014,Sopo2021DeFungi,Sadeghi2024COVID,Sadeghi2024ECG}, and disease prediction or clinical decision support \cite{Fiorini2019,Mahajan2024,Tavakolian2022FastCOVID,Tavakolian2023Readmission,Wang2022SoftwareImpacts}. These successes illustrate the robustness of deep models in handling noisy, nonlinear, and high-dimensional data, and motivate their use for complex financial series such as cryptocurrency prices.

In summary, existing approaches variously use price data, blockchain structural features, and external signals (e.g., social media, search trends), typically in a single-stream model. In contrast, our proposed PGRU method explicitly leverages parallel GRU streams for different feature sets (price and structural data), then fuses their predictions via a feedforward network. This design reduces input dimensionality, lowers computational cost, and improves accuracy relative to prior work, while being conceptually related to multi-stream architectures successfully applied in other domains \cite{Hajati2017DynamicTexture,Sadeghi2024ECG,Tavakolian2022FastCOVID}.

\section{Materials and Methods}
\label{sec:methods}

\begin{figure}[htb]
  \centering
  \includegraphics[width=0.8\linewidth]{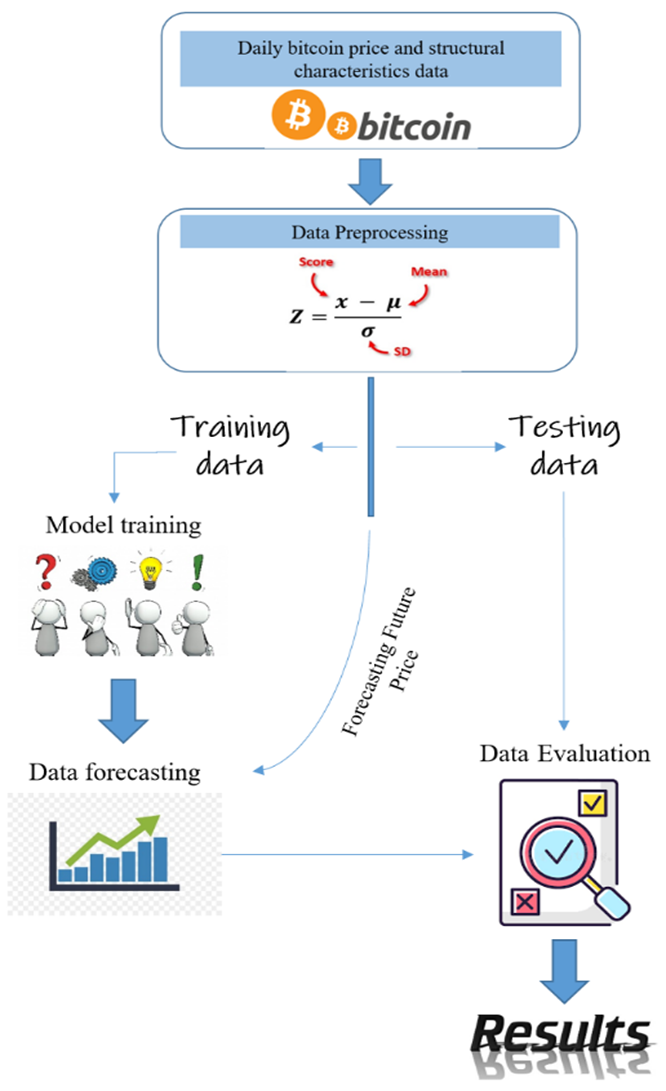}%
  \caption{Overall steps of the study: data collection, preprocessing, input--output generation, model training, and evaluation.}
  \label{fig:steps_study}
\end{figure}

In this section, we present the data sources, preprocessing steps, input--output construction, and the proposed neural network architecture for Bitcoin price prediction. Figure~\ref{fig:steps_study} summarizes the main steps.

\subsection{Data Collection}

We use reputable online sources to collect the data. Detailed structural features of Bitcoin are obtained from \cite{site_bitcoin}, and price-related features are obtained from \cite{site_investing}. For the price dataset, four daily features are used:
\begin{itemize}
    \item Average daily price,
    \item Opening price,
    \item Low price,
    \item High price.
\end{itemize}
These features help characterize daily price fluctuations. For the structural dataset, we use five blockchain and network-level features:
\begin{itemize}
    \item Average block size,
    \item Hash rate,
    \item Difficulty,
    \item Total daily transactions,
    \item Miners' revenue.
\end{itemize}

Both datasets span the period from 2016-01-01 to 2021-01-31, each containing 1,859 daily entries. A brief description of the structural features is given in Table~\ref{tab:blockchain_features}.

\begin{table}[t]
  \centering
  \caption{Bitcoin blockchain characteristics.}
  \label{tab:blockchain_features}
  \begin{tabular}{ll}
    \toprule
    Characteristic & Description \\
    \midrule
    Average Block Size & 24-hour average block size (MB) \\
    Hash Rate          & Estimated number of terahashes per second \\
    Difficulty         & Relative difficulty of finding a new block \\
    Total Transactions & Total number of daily transactions \\
    Miner Revenue      & Total value of coinbase rewards and transaction fees \\
    \bottomrule
  \end{tabular}
\end{table}

This dual-dataset design, combining price and structural features, is conceptually related to multimodal and multi-view feature integration in other domains, such as combining geometric and photometric cues for face recognition and texture matching \cite{Hajati2017Surface,Ayatollahi2015,Hajati2010PoseInvariant} or integrating clinical and imaging biomarkers in medical applications \cite{Fiorini2019,Sadeghi2024ECG}.

\subsection{Data Preprocessing}

Due to large fluctuations in Bitcoin price and structural features, the raw values span several orders of magnitude. For example, in the price dataset, the maximum average price is about 95 times greater than the minimum. In the structural dataset, the maximum difficulty is about 185 times greater than the minimum. Using simple min--max normalization would map all values into a narrow interval (e.g., $[0,1]$), causing significantly different raw values to become very close after normalization. This can hinder the network’s ability to detect meaningful fluctuations and patterns.

To address this issue, we adopt $z$-score normalization:
\[
z = \frac{x - \mu}{\sigma},
\]
where $x$ is the raw value, $\mu$ is the mean, and $\sigma$ is the standard deviation. This transformation centers the data and scales it by its standard deviation, leading to better separation between values. As a result, the network can more effectively learn the underlying patterns from the normalized data. Similar normalization strategies have been used in prior work on pattern recognition and control, where preserving relative variation is critical \cite{Hajati2017DynamicTexture,Rezaee2010FIR,Barzamini2012}.

\subsection{Input and Output Generation}

After normalization, we construct input--output pairs for supervised learning. Let $X$ and $Y$ denote the normalized price and structural datasets, respectively, each of length $n = 1859$. We define a sliding window of length $w$ (in days) to form the inputs, and use the next day’s price as the output. For the $i$-th sample, the inputs and output can be denoted conceptually as:
\[
\text{Input}_1^{(i)} = \big[ X_{i}, X_{i+1}, \dots, X_{i+w-1} \big],
\]
\[
\text{Input}_2^{(i)} = \big[ Y_{i}, Y_{i+1}, \dots, Y_{i+w-1} \big],
\]
\[
\text{Output}^{(i)} = X_{i+w},
\]
where $w$ determines the number of past days used to predict the next day's price. For example, $w=15$ means that the previous 15 days are used to predict the price on day $i+15$.

\section{Theory and Model}
\label{sec:model}

\subsection{System Model}

The proposed model consists of three neural networks:
\begin{itemize}
    \item Two parallel GRU-based recurrent networks, each responsible for capturing patterns in one input stream (price features or structural features).
    \item A final feedforward network that combines the outputs of the two GRU networks to produce the final price prediction.
\end{itemize}

Each GRU network outputs a single scalar prediction for the next-day price, based on its respective input series. The feedforward network then learns how to fuse these predictions, weighting them according to their reliability and error characteristics. This idea of parallel expert models followed by a fusion stage is analogous to multi-expert or multi-view fusion strategies used in face and texture analysis \cite{Hajati2017Surface,Hajati2017DynamicTexture} and multimodal medical AI \cite{Fiorini2019,Sadeghi2024ECG}.

\subsection{Proposed PGRU Architecture}

\begin{figure}[htb]
  \centering
  \includegraphics[width=0.9\linewidth]{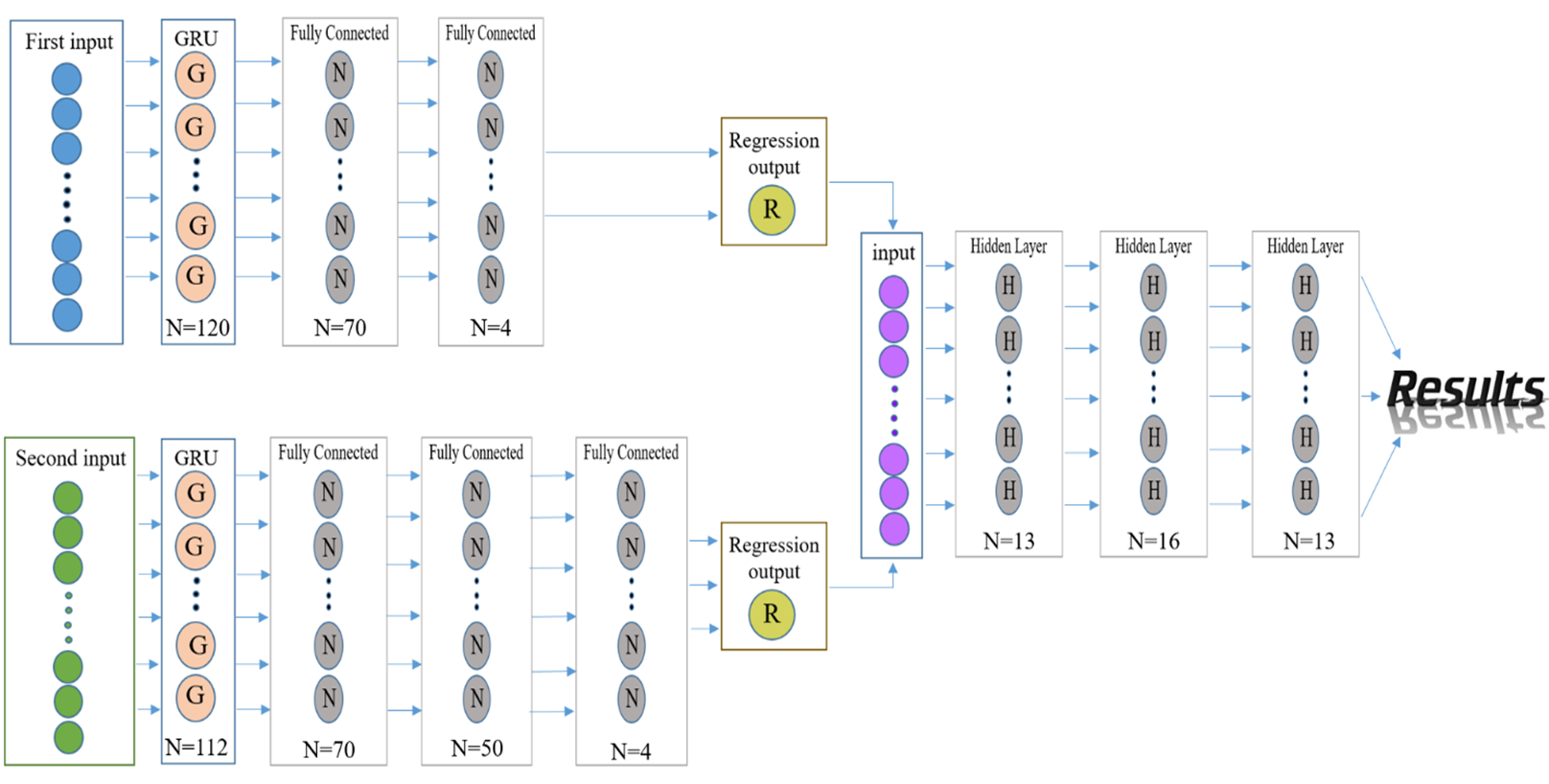}%
  \caption{Overview of the proposed PGRU architecture. Two parallel GRU-based networks process price features and structural features, respectively. Their outputs are fused by a feedforward network to produce the final price prediction.}
  \label{fig:pgru_arch}
\end{figure}

The network is implemented in MATLAB R2020b. Given the sharp fluctuations in Bitcoin prices, the architecture is designed to maximize its ability to capture temporal patterns and nonlinear relationships. Figure~\ref{fig:pgru_arch} illustrates the proposed PGRU architecture.

\paragraph{First GRU Network (Price Stream).}
The first recurrent network takes the normalized price features (average, open, low, high) over a window of length $w$ and predicts the next-day price. It consists of:
\begin{itemize}
    \item Input layer with time-series input,
    \item One or more GRU layers,
    \item Fully connected layers,
    \item Regression output layer.
\end{itemize}

\paragraph{Second GRU Network (Structural Stream).}
The second recurrent network takes the structural and network features (block size, hash rate, difficulty, total transactions, miner revenue) over the same window length $w$ and outputs its own prediction for the next-day price. Its architecture is analogous to the first network.

Both GRU networks are trained using the Adam optimizer to minimize a regression loss (e.g., mean squared error). Fully connected layers are included to improve expressiveness and reduce overfitting.

\paragraph{Fusion Network.}
The third network is a simple feedforward neural network that takes as input the two scalar predictions from the GRU networks and outputs a refined prediction. It is trained using the Levenberg--Marquardt optimizer. The fusion network implicitly learns how to weight and combine the two GRU outputs, acting as an ensemble aggregator.

Ten-fold cross-validation is applied across all three networks to improve generalization.

\subsection{Alternative LSTM-based Architecture}

\begin{figure}[htb]
  \centering
  \includegraphics[width=0.9\linewidth]{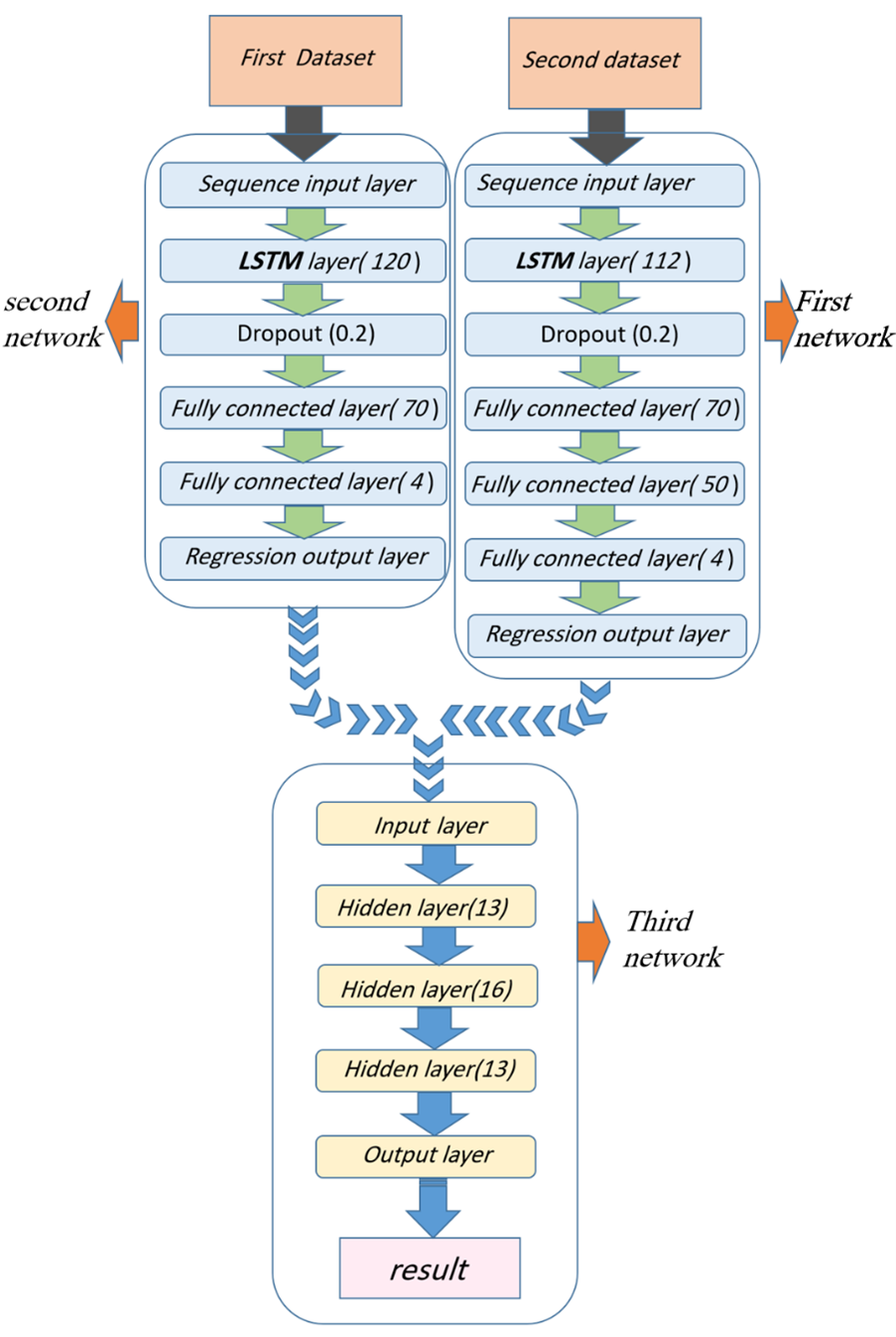}%
  \caption{Alternative architecture based on parallel LSTM networks followed by a fusion network. This structure is used as a baseline for comparison with the PGRU model.}
  \label{fig:lstm_arch}
\end{figure}

For comparison, we also design an alternative architecture in which the two parallel GRU networks are replaced by LSTM-based recurrent networks. The structure and data flow remain the same: two parallel LSTM networks (price and structural streams), followed by a feedforward fusion network. Figure~\ref{fig:lstm_arch} shows an overview of this LSTM-based architecture.

\subsection{Evaluation Metrics}

We evaluate model performance using standard regression metrics:
\begin{align}
  \text{MSE} &= \frac{1}{N} \sum_{i=1}^{N} (R_i - \hat{R}_i)^2, \\
  \text{RMSE} &= \sqrt{\text{MSE}}, \\
  \text{MAE} &= \frac{1}{N} \sum_{i=1}^{N} |R_i - \hat{R}_i|, \\
  \text{MAPE} &= \frac{100}{N} \sum_{i=1}^{N} \left| \frac{R_i - \hat{R}_i}{R_i} \right|,
\end{align}
where $R_i$ is the true price and $\hat{R}_i$ is the predicted price for sample $i$, and $N$ is the number of observations. These metrics are standard in both financial time series and other prediction tasks such as load forecasting \cite{Barzamini2012}, disease prediction \cite{Mahajan2024,Fiorini2019}, and control applications \cite{Rezaee2017MPC,Ramezani2024Drones}.

\section{Results}

\subsection{Window-Length Sensitivity}

We first evaluate the PGRU model using the full dataset (2016-01-01 to 2021-01-31) with different window lengths $w \in \{5,10,15,20,25\}$. The model is trained for 200 epochs. The first and second networks use Adam, while the fusion network uses the Levenberg--Marquardt algorithm. Ten-fold cross-validation is used throughout.

Table~\ref{tab:pgru_accuracy} summarizes the test performance for different window lengths.

\begin{table}[t]
  \centering
  \caption{Accuracy of the proposed PGRU model for different window lengths $w$.}
  \label{tab:pgru_accuracy}
  \begin{tabular}{ccccc}
    \toprule
    $w$ & MSE & RMSE & MAE & MAPE \\
    \midrule
     5  & 49{,}334.9 & 222.1 & 120.2 & 2.0\% \\
    10  & 50{,}941.0 & 225.7 & 127.2 & 2.6\% \\
    15  & 54{,}575.6 & 233.6 & 136.0 & 2.6\% \\
    20  & 61{,}648.1 & 248.3 & 143.3 & 3.2\% \\
    25  & 63{,}330.9 & 251.7 & 165.4 & 3.9\% \\
    \bottomrule
  \end{tabular}
\end{table}

\begin{figure}[htb]
  \centering
  \includegraphics[width=0.9\linewidth]{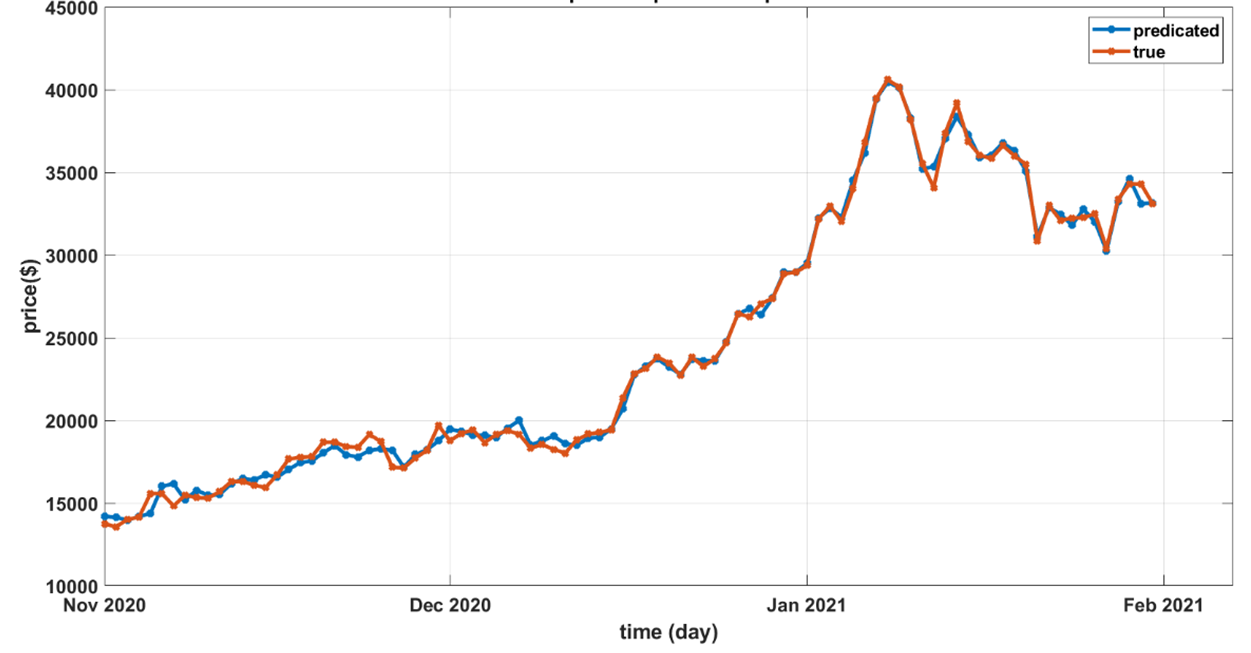}%
  \caption{Prediction result and true Bitcoin price with $w=15$ from November 1, 2020 to January 31, 2021 using the PGRU model.}
  \label{fig:pgru_w15_pred}
\end{figure}

\begin{figure}[htb]
  \centering
  \includegraphics[width=0.9\linewidth]{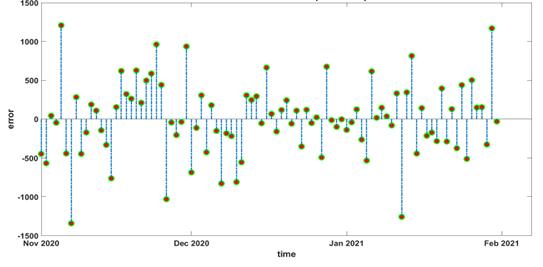}%
  \caption{Absolute differences between prediction and true price with $w=15$ for the same period as Figure~\ref{fig:pgru_w15_pred}.}
  \label{fig:pgru_w15_diff}
\end{figure}

Figures~\ref{fig:pgru_w15_pred} and \ref{fig:pgru_w15_diff} illustrate the predicted versus true prices and the absolute differences for $w=15$ over a highly volatile period (1 November 2020 to 31 January 2021).

As expected, the network tracks the overall trend and many local fluctuations, with errors ranging from small (e.g., about \$30 on some days) to larger deviations (e.g., around \$1{,}200 on certain high-volatility days). Similar behavior is observed for $w=20$.

\subsection{Multi-day Forecasting}

\begin{figure}[htb]
  \centering
  \includegraphics[width=0.9\linewidth]{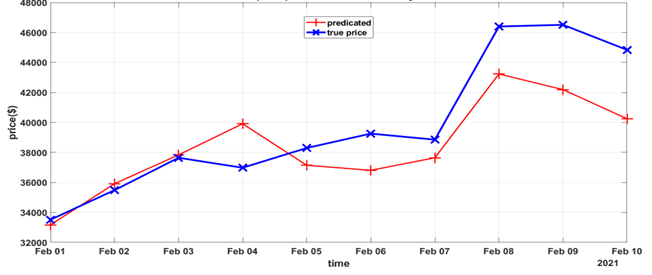}%
  \caption{Predicted and actual Bitcoin prices for a 10-day horizon using the PGRU model with $w=20$.}
  \label{fig:pgru_10day}
\end{figure}

In a second experiment, we use the PGRU model to forecast Bitcoin prices 10 days ahead, using data from January 2016 to December 2021 and $w=20$. Figure~\ref{fig:pgru_10day} shows the predicted and actual prices over this 10-day horizon. Table~\ref{tab:future_error_pgru} reports the absolute and percentage errors.

\begin{table}[t]
  \centering
  \caption{Future price prediction errors (PGRU, 10-day horizon).}
  \label{tab:future_error_pgru}
  \begin{tabular}{ccccc}
    \toprule
    Day & True Price & Predicted Price & Abs.\ Error & Abs.\ Perc.\ Error \\
    \midrule
     1 & 33{,}515.7 & 33{,}174.3 &   341.4 &  1.02\% \\
     2 & 35{,}485.2 & 35{,}913.2 &   428.0 &  1.21\% \\
     3 & 37{,}646.8 & 37{,}837.7 &   190.9 &  0.51\% \\
     4 & 36{,}982.1 & 39{,}913.6 & 2{,}931.5 &  7.93\% \\
     5 & 38{,}297.6 & 37{,}145.8 & 1{,}151.8 &  3.01\% \\
     6 & 39{,}256.6 & 36{,}810.9 & 2{,}445.7 &  6.23\% \\
     7 & 38{,}852.9 & 37{,}646.1 & 1{,}206.8 &  3.11\% \\
     8 & 46{,}395.7 & 43{,}232.2 & 3{,}163.5 &  6.82\% \\
     9 & 46{,}508.6 & 42{,}189.7 & 4{,}318.9 &  9.29\% \\
    10 & 44{,}836.0 & 40{,}245.3 & 4{,}590.7 & 10.24\% \\
    \bottomrule
  \end{tabular}
\end{table}

\subsection{LSTM-based Baseline}

We also evaluate the alternative LSTM-based model using the same datasets and window lengths $w \in \{5,10,15,20,25\}$. Table~\ref{tab:lstm_accuracy} presents the training/test performance measured by MSE, RMSE, MAE, and MAPE.

\begin{table}[t]
  \centering
  \caption{Accuracy of the LSTM-based model for different window lengths $w$.}
  \label{tab:lstm_accuracy}
  \begin{tabular}{ccccc}
    \toprule
    $w$ & MSE & RMSE & MAE & MAPE \\
    \midrule
     5  & 121{,}202.8 & 348.1 & 143.8 & 3.2\% \\
    10  & 122{,}310.3 & 349.7 & 147.6 & 3.5\% \\
    15  & 122{,}560.9 & 350.0 & 150.8 & 3.7\% \\
    20  & 133{,}676.5 & 365.6 & 168.3 & 4.1\% \\
    25  & 146{,}133.4 & 382.2 & 170.3 & 4.2\% \\
    \bottomrule
  \end{tabular}
\end{table}

\begin{figure}[htb]
  \centering
  \includegraphics[width=0.9\linewidth]{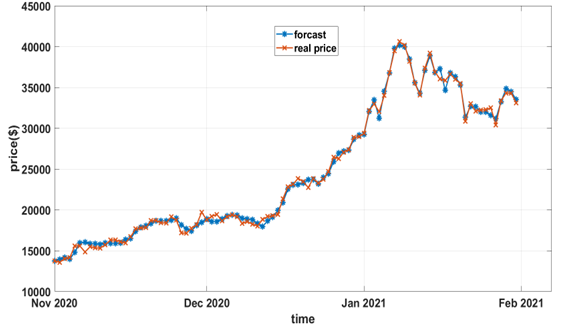}%
  \caption{Prediction result and true Bitcoin price with $w=20$ using the LSTM-based model.}
  \label{fig:lstm_w20_pred}
\end{figure}

\begin{figure}[htb]
  \centering
  \includegraphics[width=0.9\linewidth]{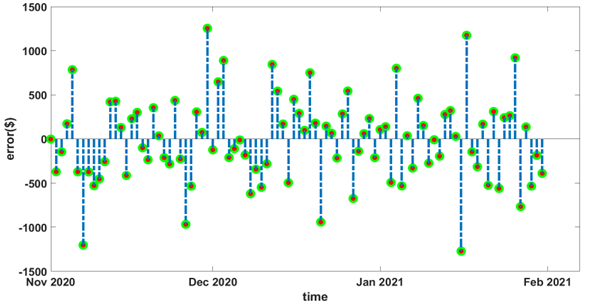}%
  \caption{Absolute differences between prediction and true price with $w=20$ for the LSTM-based model.}
  \label{fig:lstm_w20_diff}
\end{figure}

Figures~\ref{fig:lstm_w20_pred} and \ref{fig:lstm_w20_diff} show example predictions and errors for $w=20$ over the November 2020--January 2021 period.

\begin{figure}[htb]
  \centering
  \includegraphics[width=0.9\linewidth]{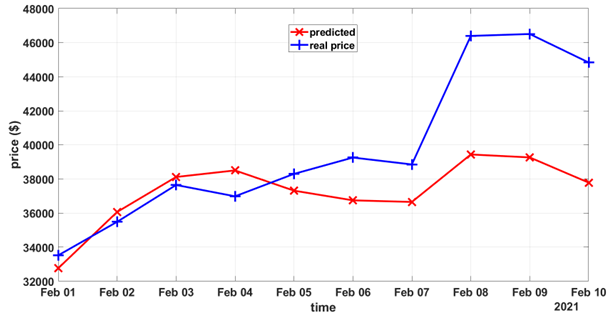}%
  \caption{Predicted and actual Bitcoin prices for a 10-day horizon using the LSTM-based model.}
  \label{fig:lstm_10day}
\end{figure}

We further used the LSTM-based network to forecast prices for 10 days ahead (February 1--10, 2021). Figure~\ref{fig:lstm_10day} shows the predicted and actual prices, and Table~\ref{tab:lstm_10day_error} reports the corresponding errors.

\begin{table}[t]
  \centering
  \caption{Future price prediction errors (LSTM-based model, 10-day horizon).}
  \label{tab:lstm_10day_error}
  \begin{tabular}{ccccc}
    \toprule
    Day & True Price & Predicted Price & Abs.\ Error & Abs.\ Perc.\ Error \\
    \midrule
     1 & 33{,}515.7 & 32{,}756.6 &   759.1 &  2.26\% \\
     2 & 35{,}485.2 & 36{,}062.5 &   577.3 &  1.62\% \\
     3 & 37{,}646.8 & 38{,}116.5 &   469.7 &  1.24\% \\
     4 & 36{,}982.1 & 38{,}502.0 & 1{,}520.0 &  4.11\% \\
     5 & 38{,}297.6 & 37{,}314.4 &   983.2 &  2.56\% \\
     6 & 39{,}256.6 & 36{,}746.4 & 2{,}511.2 &  6.39\% \\
     7 & 38{,}852.9 & 36{,}651.5 & 2{,}201.4 &  5.66\% \\
     8 & 46{,}395.7 & 39{,}437.6 & 6{,}958.1 & 14.99\% \\
     9 & 46{,}508.6 & 39{,}265.4 & 7{,}243.2 & 15.57\% \\
    10 & 44{,}836.0 & 37{,}784.1 & 7{,}051.9 & 15.72\% \\
    \bottomrule
  \end{tabular}
\end{table}

\subsection{Computational Cost}

We also compare the execution time of the GRU- and LSTM-based networks on a machine with an Intel Core i7-6700HQ CPU and an NVIDIA GTX 960 GPU. Code was carefully optimized to reduce unnecessary overhead, following best practices used in previous optimization and high-performance computing works \cite{Rezaee2014FuzzyCloud,Gavagsaz2018LoadBalancing,BarolliAINA2024,BarolliBWCCA2019,BarolliWAINA2019}. Table~\ref{tab:time_comparison} reports the average execution time (over three runs) for different window lengths.

\begin{table}[t]
  \centering
  \caption{Implementation time of GRU- and LSTM-based networks (minutes:seconds).}
  \label{tab:time_comparison}
  \begin{tabular}{cccccc}
    \toprule
    Model & $w=5$ & $w=10$ & $w=15$ & $w=20$ & $w=25$ \\
    \midrule
    GRU-based  & 10:58 & 11:37 & 12:18 & 12:56 & 13:43 \\
    LSTM-based & 12:51 & 13:29 & 14:27 & 15:03 & 15:49 \\
    \bottomrule
  \end{tabular}
\end{table}

\subsection{Comparison with Previous Work}

Table~\ref{tab:comparison_prev} compares the proposed PGRU method with several existing approaches in terms of MAPE. Our method achieves 3.24\% for $w=20$ and 2.64\% for $w=15$, outperforming time-series models, SVM- and PSO-based methods, and several LSTM/GRU-based approaches.

\begin{table}[t]
  \centering
  \caption{Comparison of the proposed PGRU method with existing research.}
  \label{tab:comparison_prev}
  \begin{tabular}{lcc}
    \toprule
    Method & Error (MAPE) & Notes \\
    \midrule
    Proposed (PGRU) & 3.24\% ($w=20$), 2.64\% ($w=15$) & This work \\
    Roy et al.\ \cite{roy2018bitcoin} & 9.69\% & Time-series forecasting \\
    Hitam et al.\ \cite{hitam2019optimized} & 9.6\% & PSO-optimized SVM \\
    Garg et al.\ \cite{garg2018arima} & $< 6\%$ & ARIMA-based \\
    Rizwan et al.\ \cite{rizwan2019bitcoin} & 5.3\% & GRU-based DL model \\
    Ji et al.\ \cite{ji2019comparative} & 4.46\% ($w=20$) & LSTM-based model \\
    Patel et al.\ \cite{patel2020deep} & 4.94\% & GRU for Litecoin \\
    \bottomrule
  \end{tabular}
\end{table}

\section{Discussion}

We introduced an innovative GRU-based parallel architecture for Bitcoin price prediction. By using two distinct input streams (price and structural features) and combining them via a fusion network, the proposed PGRU model achieves higher accuracy than prior methods while using fewer input features and incurring lower computational cost.

Our PGRU model obtains average errors of about 3.2\% for $w=20$ and 2.6\% for $w=15$. Compared to existing research, these results represent a substantial improvement, especially given the reduced feature set and smaller input dimensionality. In addition, most previous works used min--max normalization, whereas we employ $z$-score normalization, which leverages the mean and standard deviation of the data. This choice improves the separation of values and, in our experiments, increases predictive accuracy by approximately 5--7 percentage points relative to simple min--max scaling, similar to gains reported in other applications where preserving relative variation is crucial \cite{Hajati2017DynamicTexture,Barzamini2012,Rezaee2010FIR}.

Importantly, the proposed architecture is not limited to Bitcoin. It can be applied to other cryptocurrencies or economic indicators by simply replacing the input datasets. Once the relevant time-series features are collected, the network normalizes the data, learns the relationships between inputs and outputs, and predicts future values for the target index. This generality mirrors the adaptability observed in prior works where similar neural or evolutionary architectures have been successfully transferred across domains---from face and texture recognition \cite{Hajati2006FaceLocalization,Hajati2017Surface,Ayatollahi2015,AbdoliHajati2014,Shojaiee2014Palmprint,CremersACCV2014} to smart-grid forecasting and control \cite{Barzamini2012,Rezaee2017MPC,RezaeeGolpayegani2012} and medical/clinical applications \cite{Fiorini2019,Mahajan2024,Sadeghi2024ECG,Tavakolian2022FastCOVID,Tavakolian2023Readmission}.

\section{Conclusion}

Cryptocurrency price prediction remains a key challenge for economists and investors, and many methods have been proposed across economics and computer science. Artificial neural networks, especially recurrent architectures such as LSTM and GRU, are powerful tools for time-series prediction.

In this study, we presented a GRU-based neural architecture with parallel streams for Bitcoin price prediction. The model uses two categories of input data---price features and structural blockchain features---and processes them independently in two GRU networks. A feedforward fusion network then combines their outputs to produce the final prediction. The proposed method achieves MAPE of approximately 3\%, outperforming several state-of-the-art methods while requiring fewer input features and lower computational cost.

For future work, one or more additional input streams could be incorporated, including global economic indicators, prices of other major cryptocurrencies, and social-network sentiment or trend data. We expect such extensions to further improve predictive accuracy and robustness, in line with developments in multimodal AI and complex multi-agent systems \cite{BarolliAINA2024,BarolliBWCCA2019,BarolliWAINA2019,Ramezani2024Drones,Sarvghad2011ThinkingStyles}.

\bibliographystyle{unsrt}

\end{document}